# Dynamic Adaptation of LoRA Fine-Tuning for Efficient and Task-Specific Optimization of Large Language Models

Xiaoxuan Liao
New York University
New York, USA

Chihang Wang
New York University
New York, USA

Shicheng Zhou
University of Minnesota
Minneapolis, USA

Jiacheng Hu
Tulane University
New Orleans, USA

Hongye Zheng
The Chinese University of Hong Kong
Hong Kong, China

Jia Gao*
Stevens Institute of Technology
Hoboken, USA

*Abstract*—This paper presents a novel methodology of fine-tuning for large language models—dynamic LoRA. Building from the standard Low-Rank Adaptation framework, this methodology further adds dynamic adaptation mechanisms to improve efficiency and performance. The key contribution of dynamic LoRA lies within its adaptive weight allocation mechanism coupled with an input feature-based adaptive strategy. These enhancements allow for a more precise fine-tuning process that is more tailored to specific tasks. Traditional LoRA methods use static adapter settings, not considering the different importance of model layers. In contrast, dynamic LoRA introduces a mechanism that dynamically evaluates the layer's importance during fine-tuning. This evaluation enables the reallocation of adapter parameters to fit the unique demands of each individual task, which leads to better optimization results. Another gain in flexibility arises from the consideration of the input feature distribution, which helps the model generalize better when faced with complicated and diverse datasets. The joint approach boosts not only the performance over each single task but also the generalization ability of the model. The efficiency of the dynamic LoRA was validated in experiments on benchmark datasets, such as GLUE, with surprising results. More specifically, this method achieved 88.1% accuracy with an F1-score of 87.3%. Noticeably, these improvements were made at a slight increase in computational costs: only 0.1% more resources than standard LoRA. This balance between performance and efficiency positions dynamic LoRA as a practical, scalable solution for fine-tuning LLMs, especially in resource-constrained scenarios. To take it a step further, its adaptability makes it a promising foundation for much more advanced applications, including multimodal tasks.

I. INTRODUCTION

The emergence of LLMs has revolutionized the area of NLP, establishing new performance baselines for a multitude of tasks. However, the large number of parameters and complicated training procedures of such models are substantial hurdles [1]. Specializing such models to specific applications often results in prohibitively high costs and an enormous demand for computational resources. Traditional fine-tuning methods, where all model parameters are updated, aggravate these challenges due to the requirements of large computational resources and storage sizes. To deal with these challenges, parameter-efficient fine-tuning techniques have become popular, of which Low-Rank Adaptation (LoRA) is one of the most effective solutions [2].

LoRA simplifies fine-tuning by adding low-rank parameter adapters to the model, which leaves the core model parameters untouched. It saves a large amount of computation and memory resources normally required in standard fine-tuning. However, existing LoRA implementations mostly follow uniform adapter configurations for all model layers. This static configuration does not consider the varying functions and importance of distinct layers in learning tailored to specific tasks, thereby constraining the model's capacity to effectively utilize inter-layer interactions. Furthermore, in scenarios requiring complex or fluctuating input patterns, such rigid arrangements frequently prove inadequate, hindering both flexibility and overall efficacy.

Confronting the challenge of the constraints, in this work a framework of LoRA fine-tuning dynamic adaptation is developed [3]; different from most traditional methodologies, it equips a runtime mechanism where adapter parameters will effectively adapt with respect to changing demands due to the task itself and due to the importance between the various layers. Through estimation of the contribution of each layer in the overall task in an ongoing fashion, the framework can reallocate the parameters in the most effective way to enhance the utilization of each parameter. It also incorporates a data-driven approach that accommodates variations in input feature distributions [4], enhancing the model's capacity to handle complex and diverse real-world data.

The dynamic adaptation framework brings in many unique advantages. This allows for fine-tuning each layer for particular needs through precise adjustments of parameter distribution at train time [5]. In such a way, it would improve overall effectiveness in fine-tuning and raise the performance compared with the standard static LoRA setting. The flexibility of this approach strikes a very effective balance between parameter efficiency and task accuracy, hence making it practical and scalable for deploying LLMs in diverse applications [6].

Empirical studies further prove the effectiveness of dynamic LoRA. This framework outperforms the performances of standard approaches on various benchmark datasets, showing much better adaptability and generalization across diverse task environments [7]. Most importantly, these improvements do not introduce substantial additional computational and storage costs, preserving the cost-effectiveness that makes LoRA so appealing for fine-tuning. In summary, the dynamic LoRA framework constitutes a very significant step toward the fine-tuning of large language models. Through the seamless integration of dynamic parameter allocation and data-driven optimization, this provides a truly flexible, general, and powerful solution tailored toward modern NLP tasks. This work may be further explored in future directions, either by enhancing methodology for layer-importance assessment or extending this framework to multimodal [8] and resource-constrained settings, allowing for more possibility of innovation and real-world deployment.

## II. RELATED WORK

Low-rank adaptation (LoRA) has become a crucial method for parameter-efficient fine-tuning (PEFT) of large language models (LLMs). By incorporating low-rank parameter adapters into conventional LoRA frameworks, it strikes an effective balance between computational efficiency and performance. Building on this foundation, Yang et al. [9] introduced LoRA-LiteE, a fine-tuning approach tailored specifically to improve chatbot preferences. Their work highlighted LoRA's adaptability for task-specific applications and underscored the importance of carefully designed adapter configurations. However, a key limitation of these static architectures is their inability to adjust parameters dynamically—a challenge this study addresses by proposing a method for dynamic parameter reallocation based on layer importance and input distribution.

Optimization plays an especially critical role in resource-constrained settings. Wang [10] proposed a dynamic scheduling mechanism to enhance resource efficiency, laying the foundation for adjusting LoRA adapter parameters dynamically across model layers to meet task-specific needs. Similarly, Hu et al. [11] applied adaptive weight masking for few-shot learning in conditional GANs, ensuring efficient parameter utilization in low-data environments—a concept mirrored in this study's dynamic, layer-wise parameter allocation strategy. Collectively, these advancements underscore the transformative potential of dynamic methods for improving fine-tuning efficiency.

This research also draws upon broader developments in computational methodologies. Li et al. [12] investigated matrix logic techniques for identifying frequent patterns in large datasets, emphasizing the importance of computational efficiency—an insight integral to the scalability of this framework. Feng et al. [13] demonstrated the advantages of adaptive, data-driven models through collaborative optimization with ResNeXt architectures in financial data mining, a perspective reflected in the dynamic framework proposed here. Additionally, Jiang et al. [14] addressed data imbalance using generative adversarial networks (GANs), showcasing how adaptive strategies can enhance model robustness, a principle evident in this framework's ability to handle diverse datasets.

Task-specific learning strategies have also become prominent. Yao et al. [15] proposed a hierarchical graph neural network for the prediction of stock type, which adopted a multi-level adaptation mechanism to assign different importance to model layers. The hierarchical design is similar to dynamic parameter allocation in this paper, where layers are assessed regarding the importance of adapter distribution optimization. Similarly, Zhang et al. [16] developed robust graph neural networks for stability analysis in dynamic networks, showcasing how dynamic mechanisms can enhance performance for evolving input data. Their work complements the input-driven adaptation techniques proposed here. Yan et al. [17] emphasized interpretability by transforming multidimensional time series into event sequences, reinforcing the importance of adaptable frameworks when dealing with complex datasets.

Finally, improvement in computational efficiency constantly affects the field of model optimization. Li [18] emphasized feature-driven optimization methods for high-dimensional datasets, underlining the need for computational scalability—an intrinsic part of the proposed dynamic LoRA framework, which uses input-dependent methods to improve the efficiency of parameter fine-tuning.

## III. METHOD

In the traditional LoRA method, model parameters are efficiently fine-tuned through low-rank decomposition. Specifically, on the basis of freezing the original parameters $W$, a low-rank adapter consisting of two small matrices $A \in R^{d \times r}$ and $B \in R^{r \times d}$ is added. The parameter update formula is:

$$W' = W + \triangle W = W + A \cdot B$$

Among them, $r << d$ represents the size of the rank, and $\triangle W$ is the parameter update part. In traditional methods, the parameters of A and B are fixed during training and will not be dynamically adjusted according to the characteristics of the layer or task requirements, which limits their performance on different tasks.

To address this limitation, this study introduces a dynamically adaptive LoRA fine-tuning algorithm. The central innovation lies in the development of a dynamic weight allocation mechanism, enabling the parameters of A and B to be adjusted in real-time based on the specific requirements of the task. This adaptive approach allows the model to better align with task demands, improving overall performance and efficiency. The architecture of the proposed model is illustrated in Figure 1.

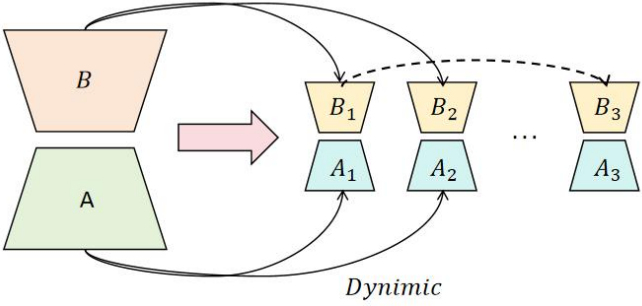

Figure 1 Overall model architecture

First, we introduce a task-related importance measurement mechanism to evaluate the importance of each layer $l$ of the model. Assuming that the loss function of the current task is $L$, the importance $\gamma_l$ of layer $l$ can be measured by calculating its contribution to the loss. The specific definition is:

$$\gamma_l = \frac{\partial L}{\partial W_l} \cdot W_l$$

This formula represents the sensitivity of the parameters to the loss function. According to the size of $\gamma_l$, we dynamically adjust the distribution weight $a_l$ of the adapter parameters so that more important layers are allocated more low-rank adaptation capabilities. The updated formula for the distribution weight is:

$$a_l = \frac{\exp(\gamma_l)}{\sum_k \exp(\gamma_k)}$$

Among them, $a_l$ represents the normalized importance weight of layer $l$, ensuring that the sum of the weights of all layers is 1.

Secondly, in the construction of the parameter adapter, we introduced a dynamic adjustment strategy based on the input feature distribution [19]. Assuming the input feature is X, the rank size $r_l$ of the adapter matrix is dynamically adjusted through the statistical information of the feature distribution (such as mean and variance). The adjustment strategy can be expressed as:

$$r_l = r_{base} \cdot (1 + \lambda \cdot Var(X_l))$$

Among them, $r_{base}$ is the base rank size, $\lambda$ is the adjustment factor, and $Var(X_l)$ represents the variance of the input feature $X_l$. This formula ensures that layers with more complex feature distributions can obtain higher adapter ranks, thereby enhancing the model's ability to capture complex features.

Next, the dynamically adjusted adapter weight $a_l$ and rank size $r_l$ are merged into the update of the low-rank adapter. The final parameter update formula is:

$$W_l' = W_l + a_l \cdot (A_l \cdot B_l)$$

Among them, $A \in R^{d \times r}$ and $B \in R^{r \times d}$ are dynamically adjusted adapter matrices.

In order to further improve the stability of the model during training, we add regularization terms to the optimization objective to balance the flexibility and stability of the adapter. The overall optimization objective is:

$$L_{total} = L_{task} + \lambda_1 \cdot \sum_l \| A_l \|_F^2 + \lambda_2 \cdot \sum_l \| B_l \|_F^2$$

Among them, $L_{task}$ is the task-related loss function, and $\lambda_1$ and $\lambda_2$ are regularization coefficients used to control the size of the adapter matrix to avoid overfitting.

In summary, this study refines the LoRA fine-tuning process by incorporating layer importance weights and input feature-driven dynamic adaptation strategies. This approach allows the model to adapt more flexibly to varying task demands while maintaining minimal computational overhead, significantly enhancing the fine-tuning performance of large language models.

IV. EXPERIMENT

*A. Datasets*

This research leverages the GLUE (General Language Understanding Evaluation) dataset as a key benchmark to evaluate the performance of the newly developed LoRA fine-tuning algorithm, which incorporates an innovative dynamic adaptation mechanism. The GLUE benchmark is a well-established and comprehensive evaluation framework in the field of natural language processing (NLP). It is specifically designed to measure a model's effectiveness across a diverse array of tasks, including text classification, sentence similarity analysis, and natural language inference. These tasks collectively reflect real-world linguistic challenges, making GLUE an indispensable tool for assessing the generalization capabilities of machine learning models.

Among the various subtasks within GLUE, the Microsoft Research Paraphrase Corpus (MRPC) and Question Natural Language Inference (QNLI) are particularly noteworthy. MRPC evaluates a model's ability to determine whether pairs of sentences share the same meaning, thereby testing its paraphrasing skills. QNLI, in contrast, examines a model's reasoning and inference abilities by requiring it to decide if a given hypothesis is supported by a corresponding premise. These complex and nuanced subtasks are critical for assessing the expressiveness and adaptability of natural language understanding systems, playing a central role in benchmarking model performance.

In our experimental design, each subtask within the GLUE dataset was treated as an independent domain for both training and testing. This structured approach ensured that evaluation results remained reliable and precise. The dataset was divided into three distinct parts: a training set for fine-tuning model parameters, a validation set for intermediate performance adjustments, and a test set for final benchmarking. This methodology aligns with established best practices in machine learning research [20], reinforcing the robustness of the evaluation process.

Through the diverse tasks provided by the GLUE dataset, this study assessed not only the overall performance of the proposed LoRA algorithm but also its capacity to adapt effectively across different linguistic challenges. This adaptability was facilitated by the dynamic adaptation mechanism embedded within the LoRA algorithm, enabling it to respond intelligently to the unique features and demands of each task.

B. Experimental Results

To thoroughly evaluate the superiority of the proposed LoRA fine-tuning algorithm, which incorporates a dynamic adaptation mechanism, this study conducted a comparative experiment against several mainstream fine-tuning methods. Specifically, the selected methods included traditional full-parameter fine-tuning, feature extraction [21] with frozen parameters, parameter-efficient adapter networks, BitFit [22] (which introduces bias parameters), and the widely adopted LoRA method [23]. Each of these approaches offers distinct characteristics, showcasing various strengths and weaknesses in terms of computational efficiency, storage requirements, and overall model performance.

This comparative framework was designed to provide a comprehensive assessment of the dynamic adaptation mechanism's impact. By examining these methods across multiple task scenarios, the experiment aimed to validate the ability of the proposed dynamic LoRA algorithm to achieve a balance between performance enhancement and resource efficiency, as well as its capacity for multi-task generalization.

Initial evaluations utilized classification experiments to verify the effectiveness of the approach. The results of these experiments, summarized in Table 1, offer a detailed comparison of performance metrics, further highlighting the advantages of the dynamic LoRA algorithm in addressing diverse linguistic tasks.

Table 1 Experimental Results

| Model | ACC | AUC | F1 | Recall |
|---|---|---|---|---|
| Full Fine-tuning | 85.3 | 88.2 | 84.1 | 83.7 |
| Feature Extraction | 81.5 | 84.0 | 80.2 | 79.8 |
| Adapter | 86.1 | 89.1 | 85.3 | 84.9 |
| BitFit | 85.8 | 88.5 | 84.7 | 84.3 |
| LORA | 87.4 | 90.2 | 86.5 | 86.0 |
| Ours | 88.1 | 91.0 | 87.3 | 86.8 |

The experimental results reveal significant variations in the performance of different fine-tuning methods across various metrics, highlighting the trade-offs between performance and efficiency. For example, Full Fine-tuning achieves an accuracy (ACC) of 85.3%. While this method demonstrates stable performance overall, it requires updating all model parameters, leading to high computational and storage overhead. As a result, it is not an ideal choice for scenarios with resource constraints, despite its relatively high baseline performance.

In contrast, Feature Extraction shows the lowest ACC at 81.5%, making it the weakest performer among all methods. This approach freezes model parameters and leverages pre-trained features for downstream tasks without adapting to task-specific information. Consequently, it struggles in more complex task scenarios and underperforms in key metrics such as AUC, F1, and Recall, limiting its ability to fully utilize the model's potential.

The Adapter and BitFit methods achieve similar results, with ACC scores of 86.1% and 85.8%, respectively. Both methods employ parameter-efficient fine-tuning strategies, either by introducing a small number of task-specific parameters (Adapter) or adjusting the model's bias parameters (BitFit). Adapter shows a slight edge in AUC and F1 scores, likely due to its stronger adaptability to specific tasks. While these methods strike a better balance between resource consumption and performance, they still lag behind LoRA and dynamic LoRA in overall effectiveness.

LoRA and the proposed dynamic LoRA (Ours) outperform all other methods across all metrics. LoRA achieves an ACC of 87.4%, while dynamic LoRA further improves to 88.1%. Notably, dynamic LoRA achieves substantial improvements in key metrics such as AUC (91.0%), F1 (87.3%), and Recall (86.8%). These results demonstrate the ability of the dynamic adaptation mechanism to capture task-specific features more effectively and optimize the adaptation process. By enhancing performance without compromising efficiency, dynamic LoRA emerges as an exceptional fine-tuning solution that balances both effectiveness and resource optimization.

Following the performance evaluation, the next step involves assessing the speed and computational overhead of the improved method to confirm its efficiency advantages. These results are detailed in Table 2.

Table 2 Speed and computational overhead experimental results

| Model | Training Time | Inference Time | Train Parameters |
|---|---|---|---|
| Full Fine-tuning | 12.5 | 15.2 | 100% |
| Feature Extraction | 6.8 | 12.7 | 0% |
| Adapter | 8.2 | 13.5 | 3% |
| BitFit | 7.9 | 13.2 | 1.5% |
| LORA | 7.1 | 12.9 | 0.8% |
| Ours | 7.4 | 13.0 | 0.9% |

The experimental results reveal significant differences in training time, inference time, and the ratios of trainable parameters across the fine-tuning methods evaluated. Full Fine-

Tuning, while delivering robust performance, proved to be the most resource-intensive approach. It required 12.5 hours of training and had the highest inference time at 15.2 milliseconds. These demanding computational requirements stem from the need to update all model parameters. While this enables exceptional adaptability and task-specific performance, the steep resource demands make it impractical for computationally constrained settings.

At the other extreme, Feature Extraction stood out for its speed and efficiency. It completed training in just 6.8 hours and achieved the lowest inference time at 12.7 milliseconds. This efficiency results from its minimalist approach, which involves training only task-specific classification heads while leaving the pre-trained model parameters untouched. However, this streamlined method comes at the expense of adaptability, offering only modest improvements in task-specific performance.

Adapter and BitFit methods represent a middle ground, balancing computational efficiency with adaptability. Adapter required 8.2 hours of training, slightly more than BitFit's 7.9 hours due to its additional parameter adaptation layer. Their inference times were also marginally higher than Feature Extraction, measured at 13.5 milliseconds for Adapter and 13.2 milliseconds for BitFit. By selectively updating a small subset of parameters, these methods achieve better flexibility without imposing significant computational costs, making them well-suited for a variety of applications.

The most notable findings, however, pertain to LoRA and the newly proposed dynamic LoRA (Ours). Both methods excel at balancing efficiency and performance, maintaining trainable parameter ratios below 1%—LoRA at 0.8% and dynamic LoRA at 0.9%. Training times were competitive, with LoRA completing in 7.1 hours and dynamic LoRA in 7.4 hours. Their inference times were nearly identical: 12.9 milliseconds for LoRA and 13.0 milliseconds for dynamic LoRA. Dynamic LoRA's adaptive mechanism allows it to achieve superior task-specific performance while introducing minimal computational overhead. This makes it particularly attractive for resource-limited environments.

Dynamic LoRA's ability to combine resource efficiency with high adaptability highlights its potential as a versatile fine-tuning solution. Its dynamic parameter adjustment mechanism enhances task-specific outcomes without compromising efficiency.

Finally, the trend shown in Figure 2, which illustrates the reduction in the loss function during fine-tuning, provides further validation for the performance and efficiency of the evaluated methods.

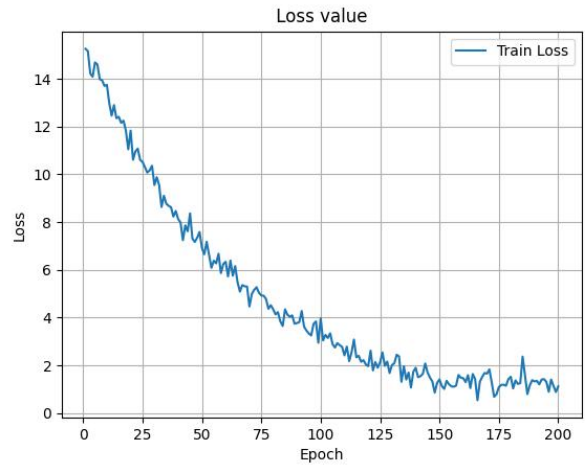

Figure 2 Loss function decline graph during fine-tuning

The downward trend in the loss function, as illustrated in Figure 2, highlights the stability of the model's training process. In the initial phases, the loss decreases sharply, reflecting the model's effective progress in learning and adapting to the data. Around the 150-epoch mark, however, the rate of loss reduction slows considerably. This marks the beginning of a convergence phase, during which the model's parameters approach their optimal configuration. Beyond this point, additional training results in only marginal improvements, with the loss exhibiting minimal further change.

In the later stages of training, the loss curve shows slight fluctuations while maintaining consistently low values overall. These minor variations are likely attributable to the inherent complexity of the data distribution. However, the limited magnitude of these fluctuations suggests that the model is not overfitting to the training data. By the end of the fine-tuning process, the loss stabilizes at approximately 2, demonstrating the model's strong performance and effective adaptation to the task.

## V. CONCLUSION

The experimental results emphasize the benefits of the dynamic adaptive LoRA fine-tuning algorithm, highlighting its clear superiority over traditional LoRA methods in a range of task scenarios. By integrating a dynamic weight allocation mechanism with an input-feature-driven adaptation strategy, the algorithm achieves significant improvements in both accuracy and efficiency. These enhancements are especially evident in tasks with high complexity and variability, showcasing the framework's ability to balance computational efficiency with exceptional performance. As a result, dynamic LoRA stands out as a highly effective and practical solution for fine-tuning large language models in challenging and diverse settings.

Although these results are promising, there is room for further improvement. Strengthening the methodology for evaluating layer importance could enable more precise weight distribution, thereby boosting the algorithm's effectiveness. Additionally, improving the dynamic adaptation strategy could better address complex variations in input features, enhancing the framework's robustness and flexibility. Aligning parameter configurations more closely with specific task requirements also offers significant opportunities for further customization and optimization.

Future research could explore expanding the algorithm's application to multimodal datasets, facilitating the integration of text and visual data for tasks that require cross-modal understanding. Another promising direction involves deploying dynamic LoRA in resource-limited environments, leveraging edge computing or distributed learning techniques to assess its performance in real-world conditions. These advancements would further improve the algorithm's practicality and accessibility, paving the way for broader adoption.

In conclusion, this study represents a major step forward in efficient fine-tuning techniques. By advancing the adaptability and scalability of the LoRA framework, it lays the groundwork for broader applications of large language models across a wide range of domains. Future work will focus on scaling the algorithm for increasingly complex tasks and exploring its potential in emerging fields, solidifying its significance as both a theoretical and practical innovation.